%% file: root.tex
\documentclass[letterpaper, 10 pt, conference]{ieeeconf}  

\usepackage{amssymb}
\usepackage{amsmath}
\usepackage{commath}
\usepackage{mathtools}

\usepackage{hyperref}
\usepackage{import}
\usepackage{paralist}
\usepackage[pdf]{svg}
\usepackage{gensymb}
\usepackage{dsfont}

\IEEEoverridecommandlockouts                              

\usepackage{graphics} 
\usepackage{epsfig} 
\usepackage{todonotes}
\usepackage[normalem]{ulem}

\title{\LARGE \bf
A Control Architecture with Online Predictive Planning for \\ Position and Torque Controlled Walking of Humanoid Robots 
}

\author{Stefano Dafarra${}^{1,2}$, Gabriele Nava${}^{1,2}$, Marie Charbonneau${}^{1,2}$, Nuno Guedelha${}^{1,2}$, Francisco Andrade${}^{1,2}$,\\ Silvio Traversaro${}^{1}$, Luca Fiorio${}^{1}$, Francesco Romano${}^{3}$, Francesco Nori${}^{3}$, Giorgio Metta${}^{1}$, and Daniele Pucci${}^{1}$
\thanks{This project has received funding from the European Union’s Horizon	2020 research and innovation programme under grant agreement No. 731540	(An.Dy).}%
\thanks{${}^{1}$ iCub Facility Department, Istituto Italiano di Tecnologia, 16163 Genova,
Italy (e-mail: name.surname@iit.it)}%
\thanks{${}^{2}$ Universit\`a degli Studi di Genova, DIBRIS}%
\thanks{${}^{3}$ Romano (fraromano@google.com) and Nori (fnori@google.com) are with Google DeepMind, London, UK.}}

\DeclareMathOperator*{\minimize}{minimize}

\begin{document}
	
\maketitle
\thispagestyle{empty}
\pagestyle{empty}

\begin{abstract}
A common approach to the generation of walking patterns for humanoid robots consists in adopting a layered control architecture. 
This paper proposes an architecture composed of three nested control loops. The outer loop exploits a robot kinematic model to plan the footstep positions. In the mid layer, a predictive controller generates a Center of Mass trajectory according to the well-known table-cart model. Through a whole-body inverse kinematics algorithm, we can define joint references for position controlled walking.
The outcomes of these two loops are then interpreted as inputs of a stack-of-task QP-based torque controller, which represents the inner loop of the presented control architecture. This resulting architecture allows the robot to walk also in torque control, guaranteeing higher level of \emph{compliance}. 
Real world experiments have been carried on the humanoid robot iCub.  
\end{abstract}

\import{tex/}{Introduction}

\import{tex/}{Background}

\import{tex/}{Architecture}
\import{tex/}{Experiments}
\import{tex/}{Conclusions}

\addtolength{\textheight}{0cm}   

\addcontentsline{toc}{section}{References}

\bibliography{IEEEabrv,Bibliography}

\end{document}

%% file: tex/Introduction.tex
\section{Introduction}
Despite decades of research in the subject, \emph{robust} bipedal locomotion of humanoid robots  is still a challenge for the Robotics community. 
The unpredictability of the terrain, the nonlinearity of the robot-environment models, and the \emph{low efficiency} of the robot actuators - that are a far cry from the human musculoskeletal system - are only a few complexities that render robot bipedal locomotion an active research domain. In this context, feedback control algorithms for robust bipedal locomotion are of primary importance. This paper contributes towards this direction by presenting an on-line predictive kinematic planner for foot-step positioning and center-of-mass (CoM) trajectories. This planner is also integrated with a stack-of-task torque controller, which ensures a degree of \emph{robot compliance} and further increases the overall system robustness to external perturbations.
 
A recent approach for bipedal locomotion control that became popular during the DARPA Robotics Challenge~\cite{feng2015optimization}
consists in defining a hierarchical  control architecture. Each layer of this architecture receives inputs from the robot and its surrounding environment, and provides references to the the layer below. The lower the layer, the shorter the time horizon that is used to evaluate the outputs. Also, lower layers usually employ more complex models to evaluate the outputs, but the shorter time horizon often results in faster computations for obtaining these outputs.

From higher to lower layers, \emph{trajectory optimization for foot-step planning}, \emph{receding horizon control (RHC)}, and \emph{whole-body quadratic programming control} represent a common stratification of the control architecture for  bipedal locomotion control.

\emph{Trajectory optimization for foot-step planning} is in charge of finding a sequence of robot footholds, which is also fundamental for maintaining the robot balance. This optimisation can be achieved by considering both kinematic and dynamic robot models \cite{dai2014whole,herzog2015trajectory,carpentier2016versatile}, and with different optimisation techniques, such as the Mixed Integer Programming~\cite{deits2014footstep}. The robust-and-repeatable application of these sophisticated techniques, however, still require time-consuming gain tuning since they usually consider complex models characterizing  the hazardous environment surrounding the robot.


For a certain number of applications, the terrain can be considered to be flat.
In these cases, it is known that the human upper body is usually kept tangent to the \emph{walking path}~\cite{flavigne2010reactive,mombaur2010human} all the more so because stepping aside, i.e. perpendicular to the path, is energetically inefficient~\cite{handford2014sideways}. All these considerations suggest to use a simple kinematic model to generate the walking trajectories: the unicycle model (see, e.g., \cite{PascalHandbook}). This model can be used to  plan footsteps in a corridor with turns and junctions using cameras~\cite{faragasso2013vision}, or to perform \emph{evasive} robot motions~\cite{cognetti2016real}. In all these cases, however,  the robot velocity is kept to a constant value.


\emph{Receding horizon control (RHC)}, also referred to as Model Predictive Control (MPC)~\cite{Mayne2000Stability}, is often in charge of finding feasible trajectories for the robot center of mass along the walking path. The computational burden to find feasibility regions, however, usually calls for using simplified models to characterise the robot dynamics. The Linear Inverted Pendulum \cite{Kajita2001} and the Capture Point \cite{Pratt2006} models represent two widespread simplified robot models. These simplified linear models
have allowed on-line RHC, also providing  references for the footstep locations \cite{diedam2008online,missura2014balanced,bombile2017capture}. 




\emph{Whole-body quadratic-programming (QP) controllers} are instantaneous algorithms that usually find (desired) joint torques and contact forces achieving some desired robot accelerations. In this framework, the generated joint-torques and contact forces can satisfy inequality constraints, which allow to meet friction constraints. The desired accelerations, that QP controllers track, shall ensure the stabilisation of reference positions coming from the RHC layer. Although  the reference positions may be  stabilised  by a  joint-position control loop, joint-torque based controllers have shown to ensure a degree of compliance, which also allows safe interactions with the environment \cite{Saab2013, Ott2011}. From the modeling point of view, full-body \emph{floating-base} models are usually employed in QP controllers. These controllers are often composed of several tasks, organised with strict or weighted hierarchies~\cite{Stephens2010, Herzog2014, lee2012, Nava2016, highlyDynamic}.

This paper presents a \emph{reactive} control architecture for bipedal robot locomotion, namely, the entire architecture uses on-line feedback from the robot and user-desired quantities. This architecture implements the above three layers, and can be launched on both position and torque controlled robots. 

The \emph{trajectory optimization for foot-step planning} is achieved by a planning module that uses a simplified kinematic robot model, namely, the unicycle model. Feet positions are updated depending on the robot state and on a desired direction coming from a joypad, which gives a human user teleoperating the robot the possibility of defining  desired walking paths.   Differently from~\cite{faragasso2013vision}, we do not fix the robot velocity. Compared to \cite{diedam2008online,missura2014balanced,bombile2017capture}, we do not assume the robot to be always in single support. As a consequence, the robot  avoids to step in place continuously if the desired robot position does not change, or changes slowly. Another drawback of these approaches is that  feet rotations are  planned separately from  linear positions, and this drawback is overcome by our approach. 

Once  footsteps are defined,  a \emph{receding horizon control (RHC)} module generates kinematically feasible trajectories for the robot center of mass and joint trajectories by using the LIP model \cite{Kajita2003} and  whole-body inverse kinematics. Hence, we separate the generation of footsteps from the computation of feasible CoM trajectories. The implemented RHC module runs at a frequency of 100 Hz.

The CoM and feet trajectories generated by the RHC module are then streamed as desired values to either a joint-position control loop, or to a \emph{whole-body quadratic-programming (QP) controller} running at 100 Hz. This latter controller generates desired joint torques, ensuring a degree of robustness and robot compliance. The desired joint torques are then stabilised by a low-level joint torque controller running at 1 kHz. Experimental validations of the proposed approach are carried out on the iCub humanoid robot \cite{Nataleeaaq1026}, with both position and joint torque control experiments.

In light of the above, this paper presents a torque-control, on-line RHC architecture for bipedal robot locomotion. 



%% file: tex/Background.tex
\section{Background}

\subsection{Notation}
In this paper we are going to use the following notation:
\begin{itemize}
	\item The $i_{th}$ component of a vector $x$ is denoted as $x_i$.	
	\item $\mathcal{I}$ is a fixed inertial frame with respect to (w.r.t.) 
	which the robot's absolute pose is measured.
	\item $e_1 := (1,0)^T$ and $e_2 :=(0,1)^T$ denote the canonical basis vectors of $\mathbb{R}^2$.
	\item $R_2(\theta) \in SO(2)$ is the rotation matrix of an angle $\theta \in \mathbb{R}$; $S_2=R_2(\pi/2)$ is the unitary skew-symmetric matrix.
	\item given a function of time $f(t)$ the dot notation denotes the time derivative, i.e.
	$\dot{f} := \frac{\dif f}{\dif t}$. Higher order derivatives are denoted with a corresponding amount of dots.
	\item $\mathds{1}_n \in \mathbb{R}^{n \times n}$ denotes the identity matrix of dimension $n$.
	\item $S(\cdot)$ is the skew-symmetric operation associated with the cross product in $\mathbb{R}^3$.
	\item The vee operator denoted by $(\cdot)^{\vee}$ is the inverse of the $S(\cdot)$ operation. 
	\item The Euclidean norm of a vector of coordinates $v \in \mathbb{R}^n$ is denoted by $\|v\|$.
	\item $^{A}R_{B} \in SO(3)$ and $^{A}T_{B} \in SE(3)$ denote the rotation and transformation matrices which transform a vector expressed in the $B$ frame into a vector expressed in the $A$ frame.
	\end{itemize}

\begin{figure}[tpb]
	\def\svgwidth{0.8\columnwidth}
	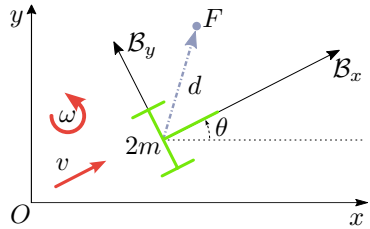
	\centering
	\caption{Notation. The unicycle model is a planar model of a robot having two wheels placed at a distance $2m$, $m \in \mathbb{R}$ with a coinciding rotation axis.  Hence, this mobile robot cannot move sideways, i.e. along the wheel axis, but it can turn by moving the wheel at different velocity. $\mathcal{B}$ is a frame attached to the robot whose origin is located in the middle of the wheels axis. Point $F$ is attached to the robot. Its position expressed in $\mathcal{B}$ is given by $d\in \mathbb{R}^2$, a constant vector.}
	\label{fig:notation}
\end{figure}

\subsection{The unicycle model}
\label{sec:unicycleModel}
The unicycle model, represented in Fig. \ref{fig:notation}. is described by the following model equations \cite{pucci2013nonlinear}:
\begin{IEEEeqnarray}{RCL}
	\IEEEyesnumber \phantomsection \label{unicycleDynamics}
	\dot x & = & v R_2(\theta) e_1, \IEEEyessubnumber \label{honolonimicCon} \\
	\dot \theta & = & \omega,       \IEEEyessubnumber
\end{IEEEeqnarray}
with $v \in \mathbb{R}$ and $\omega \in \mathbb{R}$ the robot's rolling and rotational velocity, respectively. $x \in \mathbb{R}^2$ is the unicycle position in the inertial frame $\mathcal{I}$, while $\theta \in \mathbb{R}$ represents the angle around the $z-$axis of $\mathcal{I}$ which aligns the inertial reference frame with a unicycle fixed frame.
The variables $v$ and $\omega$ are considered 
as kinematic control inputs.

A reasonable control objective for this kind of model is to asymptotically stabilize the point $F$ about a desired point $F^*$ whose position is defined as $x_F^*$. For this purpose, 
 define the error $\tilde{x}$ as 
\begin{equation}
\tilde{x}  := x_F - x_F^*. \label{eq:xtilde}
\end{equation}
so that the control objective is equivalent to the asymptotic stabilization of $\tilde{x}$ to zero.
Since
\[ x_F = x + R_2(\theta)d,\]
then by differentiation it yields 
\begin{equation}
\label{errorDynamics}
	\dot{x}_F = \dot{x} + \omega R_2(\theta) S_2 d. 
\end{equation}

Eq. \eqref{errorDynamics} describes the output dynamics. Substituting Eq. \eqref{unicycleDynamics} into Eq. \eqref{errorDynamics}, we can rewrite the output dynamics as:
\begin{equation}
\dot{x}_F = B(\theta)u,
\end{equation}
where $u = \left[v ~~ \omega\right]$ is the vector of control inputs.
It is easy to show that $\det B(\theta) = d_1$, which means that when the control point $F$ is not located on the wheels' axis, its stabilization to an arbitrary reference position $F^*$ can be achieved by the use of simple feedback laws. For example, if we define
\begin{equation}
u = B(\theta)^{-1}(\dot{x}^*_F - K\tilde{x})
\end{equation}  
with $K$ a positive definite matrix, then we have
\begin{equation} 
\dot{\tilde{x}} = -K\tilde{x}.
\end{equation}
Thus, the origin of the error dynamics is an asymptotically stable equilibrium.
Notice that this control law is not defined when $d_1=0$.

\subsection{Zero Moment Point Preview Control}
\label{sec:zmppreview}
The MPC controller implemented in this work has been derived from the Zero Moment Point (ZMP) preview control described in \cite{Kajita2003}.
This algorithm adopts a simplified model, viz. the table-cart model, based on the Linear Inverted Pendulum (LIP) approximate model.
The humanoid model is assimilated to an inverted pendulum that is linearised around the vertical position.
The ZMP can be related to the CoM position and acceleration by the following equation:
\begin{equation}
\label{eq:cop}
x_{\text{ZMP}} = P_{2D} x_{\text{CoM}} - \frac{x_{\text{CoM},z}}{g} P_{2D} \ddot{x}_\text{CoM}
\end{equation}
where $g$ is the gravitational constant, $x_\text{ZMP} \in \mathbb{R}^2$ denotes the Zero Moment Point (ZMP) \cite{vukobratovic2004zero} coordinate, while $x_\text{CoM} \in \mathbb{R}^3$ is the center of mass 3D coordinate. $x_{\text{CoM},z}$, i.e. the CoM height, is assumed constant and $P_{2D}$
is the matrix projecting the CoM on the 2D plane, i.e. it considers only the $x$ and $y$ coordinates of the center of mass.

Assuming to have a desired ZMP trajectory $x_\text{ZMP}^*(t)$, we want to track this signal at every time instant.
One possibility is to consider the ZMP as an output of following the dynamical system:
\begin{equation}
\label{eq:cart_table_dyn_ss}
\dot{\chi} = A_{\text{ZMP}} \chi + B_{\text{ZMP}} u, \quad y = C_{\text{ZMP}} \chi
\end{equation}
where the new state variable $\chi$ and control $u$ are defined as
\begin{equation}
\label{eq:cart_table_dyn_ss_statedef}
\chi := \begin{bmatrix}
x_{\text{CoM},x,y} \\
\dot{x}_{\text{CoM},x,y} \\
\ddot{x}_{\text{CoM},x,y}
\end{bmatrix} \in \mathbb{R}^6, \quad
u := \begin{bmatrix}
\dddot{x}_{\text{CoM},x,y}
\end{bmatrix} \in \mathbb{R}^2.
\end{equation}
The system matrices are defined as in the following:
\begin{equation}
\label{eq:cart_table_dyn_ss_matrices}
\begin{aligned}
A_{\text{ZMP}} & = \begin{bmatrix}
0_{2\times 2} & \mathds{1}_2 & 0_{2\times 2}\\
0_{2\times 2} & 0_{2\times 2} & \mathds{1}_2\\
0_{2\times 2} & 0_{2\times 2} & 0_{2\times 2}
\end{bmatrix}, \quad
B_{\text{ZMP}} = \begin{bmatrix}
0_{4\times 2} \\
\mathds{1}_2
\end{bmatrix} \\
C_{\text{ZMP}}& = \begin{bmatrix}
\mathds{1}_2 & 0_{2\times 2} & -\frac{x_{\text{CoM},z}}{g}\mathds{1}_2
\end{bmatrix}.
\end{aligned}
\end{equation}

Defining a cost function $\mathcal{J}$
\begin{equation}
\label{eq:cost_linear}
\mathcal{J} = \int_0^{t_f} \norm{x_\text{ZMP}^*- C_{\text{ZMP}} \chi}_Q^2  + \norm{u}^2_R \dif \tau .
\end{equation}
where $Q$ and $R$ are two weight matrices of suitable dimension. $\mathcal{J}$ penalizes the output tracking error plus a regularization term on the effort. In order to minimize $\mathcal{J}$ given Eq. \eqref{eq:cart_table_dyn_ss} we can implement a Linear Quadratic controller. This simple controller allows to track a desired ZMP trajectory.

\subsection{System Modeling}
\label{sec:modeling}
The full model of a humanoid robot is composed of $n+1$ rigid bodies, called links, connected by $n$ joints with one degree of freedom each. We also assume that none of the links has an \emph{a priori} constant pose with respect to an inertial frame, i.e. the system is \emph{free floating}.

The robot configuration space can then be characterized by the \emph{position} and the \emph{orientation} of a frame attached to a robot's link, called \emph{base frame} $\mathcal{B}$, and the joint configurations. Thus, the robot configuration space is the group $\mathbb{Q} = \mathbb{R}^3 \times SO{(3)} \times \mathbb{R}^n$. An element $q \in \mathbb{Q}$ can be defined as the following triplet: $q = (\prescript{\mathcal{I}}{}p_{\mathcal{B}}, \prescript{\mathcal{I}}{}R_{\mathcal{B}}, s)$ 
where $\prescript{\mathcal{I}}{}p_{\mathcal{B}} \in \mathbb{R}^3$ denotes the position of the base frame with respect to the inertial frame, $\prescript{\mathcal{I}}{}R_{\mathcal{B}} \in \mathbb{R}^{3\times3}$ is a rotation matrix representing the orientation of the \emph{base frame}, and $s \in \mathbb{R}^n$ is the joint configuration. 

The \emph{velocity} of the multi-body system can be characterized by the \emph{algebra} $\mathbb{V}$ of $\mathbb{Q}$ defined by: $\mathbb{V} = \mathbb{R}^3 \times \mathbb{R}^3 \times \mathbb{R}^n$.
An element of $\mathbb{V}$ is then a triplet $\nu = ( ^\mathcal{I}\dot{ p}_{\mathcal{B}},^\mathcal{I}\omega_{\mathcal{B}},\dot{s}) = (\text{v}_{\mathcal{B}}, \dot{s})$, where $^\mathcal{I}\omega_{\mathcal{B}}$ is the angular velocity of the base frame expressed w.r.t. the inertial frame, i.e. $^\mathcal{I}\dot{R}_{\mathcal{B}} = S(^\mathcal{I}\omega_{\mathcal{B}})^\mathcal{I}{R}_{\mathcal{B}}$. A more detailed description of the floating base model is provided in \cite{traversaro2017}.

We also assume that the robot is interacting with the environment exchanging $n_c$ distinct wrenches\footnote{As an abuse of notation, we define as \emph{wrench} a quantity that is not the dual of a 
\emph{twist}, but a 6D force/moment vector.}. The application of  the Euler-Poincar\'e formalism \cite[Ch. 13.5]{Marsden2010}:
\begin{align}
\label{eq:system}
{M}(q)\dot{{\nu}} + {C}(q, {\nu}){\nu} + {G}(q) =  B \tau + \sum_{k = 1}^{n_c} {J}^\top_{\mathcal{C}_k} f_k
\end{align}
where ${M} \in \mathbb{R}^{n+6 \times n+6}$ is the mass matrix, ${C} \in \mathbb{R}^{(n+6) \times (n+6)}$ accounts for Coriolis and centrifugal effects, ${G} \in \mathbb{R}^{n+6}$ is the gravity term, $B = (0_{n\times 6} , \mathds{1}_n)^\top$ is a selector matrix, $\tau \in \mathbb{R}^{n}$ is a vector representing the internal actuation torques, and $f_k \in \mathbb{R}^{6}$ denotes the $k$-th external wrench applied by the environment on the robot. The Jacobian ${J}_{\mathcal{C}_k} = {J}_{\mathcal{C}_k}(q)$ is the map between the robot's velocity ${\nu}$ and the linear and angular velocity at the $k$-th contact link.

                           
Lastly, it is assumed that a set of holonomic constraints acts on System \eqref{eq:system}. These holonomic constraints are of the form $c(q) = 0$, and may represent,
for instance, a frame having a constant pose w.r.t. the inertial frame.
In the case this frame corresponds to the location at which a rigid contact occurs on a link, we represent the holonomic constraint as ${J}_{\mathcal{C}_k}(q) {\nu} = 0.$

Hence, the holonomic constraints associated with all the rigid contacts can be represented as
\begin{IEEEeqnarray}{RCL}
\label{eqn:constraintsAll}
{J}(q) {\nu} {=} 
\begin{bmatrix}{J}_{\mathcal{C}_1}(q) \\ \cdots \\ {J}_{\mathcal{C}_{n_c}}(q)  \end{bmatrix}{\nu}  {=} 
\begin{bmatrix} J_b & J_j  \end{bmatrix}{\nu}  
&=& J_b {\text{v}}_{\mathcal{B}}+ J_j \dot{s} = 0,
\IEEEeqnarraynumspace
\end{IEEEeqnarray}
with $J_b \in \mathbb{R}^{6n_c \times 6},J_j \in \mathbb{R}^{6n_c \times n} $.
The base frame velocity is denoted by $\text{v}_{\mathcal{B}} \in \mathbb{R}^6$.

%

%% file: figures/notation_simple.pdf_tex
\begingroup%
  \makeatletter%
  \providecommand\color[2][]{%
    \errmessage{(Inkscape) Color is used for the text in Inkscape, but the package 'color.sty' is not loaded}%
    \renewcommand\color[2][]{}%
  }%
  \providecommand\transparent[1]{%
    \errmessage{(Inkscape) Transparency is used (non-zero) for the text in Inkscape, but the package 'transparent.sty' is not loaded}%
    \renewcommand\transparent[1]{}%
  }%
  \providecommand\rotatebox[2]{#2}%
  \ifx\svgwidth\undefined%
    \setlength{\unitlength}{221.69349848bp}%
    \ifx\svgscale\undefined%
      \relax%
    \else%
      \setlength{\unitlength}{\unitlength * \real{\svgscale}}%
    \fi%
  \else%
    \setlength{\unitlength}{\svgwidth}%
  \fi%
  \global\let\svgwidth\undefined%
  \global\let\svgscale\undefined%
  \makeatother%
  \begin{picture}(1,0.4501774)%
    \put(-0.00147953,0.06414063){\color[rgb]{0,0,0}\makebox(0,0)[lt]{\begin{minipage}{0.2458349\unitlength}\raggedright $O$\end{minipage}}}%
    \put(0.65068451,0.04880414){\color[rgb]{0,0,0}\makebox(0,0)[lt]{\begin{minipage}{0.19102207\unitlength}\raggedright $x$\end{minipage}}}%
    \put(0,0){\includegraphics[width=\unitlength,page=1]{notation_simple.pdf}}%
    \put(-0.00253416,0.43851856){\color[rgb]{0,0,0}\makebox(0,0)[lt]{\begin{minipage}{0.21118017\unitlength}\raggedright $y$\end{minipage}}}%
    \put(0,0){\includegraphics[width=\unitlength,page=2]{notation_simple.pdf}}%
    \put(0.08602868,0.15345313){\color[rgb]{0,0,0}\makebox(0,0)[lt]{\begin{minipage}{0.11096401\unitlength}\raggedright $v$\end{minipage}}}%
    \put(0,0){\includegraphics[width=\unitlength,page=3]{notation_simple.pdf}}%
    \put(0.0923437,0.24682529){\color[rgb]{0,0,0}\makebox(0,0)[lt]{\begin{minipage}{0.28643149\unitlength}\raggedright $\omega$\end{minipage}}}%
    \put(0,0){\includegraphics[width=\unitlength,page=4]{notation_simple.pdf}}%
    \put(0.39275847,0.23193987){\color[rgb]{0,0,0}\makebox(0,0)[lt]{\begin{minipage}{0.21471085\unitlength}\raggedright $\theta$\end{minipage}}}%
    \put(0.61919722,0.34651247){\color[rgb]{0,0,0}\makebox(0,0)[lt]{\begin{minipage}{0.39333585\unitlength}\raggedright $\mathcal{B}_x$\end{minipage}}}%
    \put(0.22360601,0.39522838){\color[rgb]{0,0,0}\makebox(0,0)[lt]{\begin{minipage}{0.38837404\unitlength}\raggedright $\mathcal{B}_y$\end{minipage}}}%
    \put(0.34043397,0.31448627){\color[rgb]{0,0,0}\makebox(0,0)[lt]{\begin{minipage}{0.15291382\unitlength}\raggedright $d$\end{minipage}}}%
    \put(0.36840051,0.44213999){\color[rgb]{0,0,0}\makebox(0,0)[lt]{\begin{minipage}{0.17546748\unitlength}\raggedright $F$\end{minipage}}}%
    \put(0,0){\includegraphics[width=\unitlength,page=5]{notation_simple.pdf}}%
    \put(0.2194322,0.19541362){\color[rgb]{0,0,0}\makebox(0,0)[lt]{\begin{minipage}{0.24291509\unitlength}\raggedright $2m$\end{minipage}}}%
  \end{picture}%
\endgroup%

%% file: tex/Architecture.tex
\section{Architecture}
In this section we summarise the components constituting the presented architecture, namely:
%
%
\begin{itemize}
	\item the footstep planner,
	\item the RH controller,
	\item the stack-of-task-balancing controller.
\end{itemize}

\subsection{The Footstep Planner}
\label{sec:footstepPlanner}
Consider the unicycle model presented in Section \ref{sec:unicycleModel}. Assuming to know the reference trajectory for point $F$ up to time $T$, we can integrate the closed-loop system described in Section \ref{sec:unicycleModel} to obtain the trajectory spanned by the unicycle. The next step consists in discretizing the unicycle trajectories at a fixed rate $1/\mathrm{d}t$. This passage allows us to search for the best foot placement option in a smaller space, constituted by the set of discretization points taken from the original unicycle trajectories. 

Given a discrete instant $k \in \mathbb{N}$, we can define $x_k = x(k\mathrm{d}t)$ as the position of the unicycle at instant $k$, while $\theta_k = \theta(k\mathrm{d}t)$ is its orientation along the $z-$axis. We can use the discretized trajectories to reconstruct the desired position of the left and right foot, $x_k^l$ and $x_k^r$, using the following relations:
\begin{IEEEeqnarray}{C}
	\IEEEyesnumber \phantomsection \label{sampledPositions}
		x_k^l = x_k + R_2(\theta_k)\begin{bmatrix} 0 \\ m \end{bmatrix},{}{} x_k^r = x_k + R_2(\theta_k)\begin{bmatrix} 0 \\ -m \end{bmatrix}. \IEEEyessubnumber\\
	\theta_k^l = \theta_k^r = \theta_k. \label{sampledAngles} \IEEEyessubnumber
\end{IEEEeqnarray}

A step contains two phases: double support and single support. During double support, both robot feet are on ground and depending on the foot, we can distinguish two different states: \textit{switch-in} if the foot is being loaded, \textit{switch-out} otherwise. In single support instead, a foot is in a \textit{swing} state if it is moving, \textit{stance} otherwise. The instant in which a foot lands on the ground is called \textit{impact time}, $t_\text{imp}^f$. After this event, the foot will experience the following sequence of states: 
$
\textit{switch-in} \rightarrow \textit{stance} \rightarrow \textit{switch-out} \rightarrow \textit{swing}.
$
This sequence is terminated by an another \textit{impact time}. At the beginning of the \textit{switch-out} phase, the \emph{other} foot has landed on the ground with an impact time $t_\text{imp}^{\sim f}$. The step duration, $\Delta_t$ is then defined as:
\begin{equation*}
\Delta_t = t_\text{imp}^{\sim f} - t_\text{imp}^f.
\end{equation*}
We define additional quantities relating the two feet when in double support ($ds$):
\begin{itemize}
	\item the orientation difference when in double support $\Delta_\theta = |\theta^l_{ds} - \theta^r_{ds}|$.
	\item The feet distance $\Delta_x = \|x^l_{ds} - x^r_{ds}\|$.
	\item The position of the left foot expressed on the frame rigidly attached to the right foot, ${}^rP_l$.
\end{itemize}
These quantities will be used to determine the feet trajectories starting from the unicycle ones.

Given the discretized unicycle trajectories, a possible policy consists in fixing the duration of a step, namely $\Delta_t = \text{const}$, or in fixing its length, i.e. setting $\Delta_x$ to a constant. Both these two strategies are not desirable. In the former case the robot will always take (eventually) very short steps at maximum speed. In the latter, if the unicycle is advancing slowly (because of slow moving references), the robot will take steps always at maximum length sacrificing the walking speed. In view of these considerations, we would like the planner to modify step length and speed depending on the reference trajectory. Since we would like to avoid fixing any variable, it is necessary to define a cost function.
It is composed of two parts. The first part weights the squared inverse of $\Delta_t$, thus penalizing fast steps, while the second penalize the squared 2-norm of $\Delta_x$, avoiding to take long steps. Thus, the cost function $\phi$ can be written as:
\begin{equation} \label{cost_fcn}
\phi = k_t \frac{1}{\Delta_t^2} + k_x \|\Delta_x\|^2,
\end{equation}
where $k_t$ and $k_x$ are two positive numbers. Depending on their ratio, the robot will take long-and-slow or short-and-fast steps.
Notice that $\phi$ is not defined when $\Delta_t = 0$, but the robot would not be able to take steps so fast. Thus, we need to bound $\Delta_t$:
\begin{equation}
	t_{min} \leq \Delta_t \leq t_{max}
\end{equation}
where the upper bound avoids a step to be too slow.

Regarding $\Delta_x$ it needs to be lower than an upper-bound $d_{max}$, bounding the swinging foot into a circular area drawn around the stance foot with radius $d_{max}$.
Another constraint to be considered is the relative rotation of the two feet. In particular the absolute value of $\Delta_\theta$ must be lower than $\theta_{max}$.

Finally, in order to avoid the robot to twist its legs, we simply resort to a bound on the $y-$component of ${}^rP_l$ vector. Indeed, it corresponds to the width of the step. By applying a lower-bound $w_{min}$ on it, we avoid the left foot to be on the right of the other foot.

Finally we can write the constraints defined above, together with $\phi$, as an optimization problem:
\begin{IEEEeqnarray}{LRCCCL}	
	\IEEEyesnumber \phantomsection \label{unicycle_optimization}
	\minimize_{t_{imp}}    & \IEEEeqnarraymulticol{5}{C}{ K_t \frac{1}{\Delta_t^2}  +  K_x \|\Delta_x\|^2 }\IEEEyessubnumber\\
	\text{s.t.}: & t_{min} & \leq & \Delta_t &\leq& t_{max} \IEEEyessubnumber \label{timeConstr}\\
	&&& \Delta_x &\leq & d_{max} \IEEEyessubnumber\\
	&&& |\Delta_\theta| &\leq&  \theta_{max} \IEEEyessubnumber \\
	& w_{min} &\leq& {}^rP_{l,y}.&& \IEEEyessubnumber
\end{IEEEeqnarray} 
The decision variables are the \textit{impact times}. If we select an instant $k$ to be the \textit{impact time} for a foot, then we can obtain the corresponding foot position and orientation using Eq. \eqref{sampledPositions} and Eq. \eqref{sampledAngles}. 

The solution of the optimization problem is obtained through a simple iterative algorithm. The initialization can be done by using the measured position of the feet. Starting from this configuration we can integrate the unicycle trajectories assuming to know the reference trajectories. Once we discretize them, we can iterate over $k$ until we find a set of points which satisfy the conditions defined above. Among the remaining points we can evaluate $\phi$ to find our optimum.

Once we planned footsteps for the full time horizon, we can proceed in interpolating the feet trajectories and other relevant quantities for the walking motion, e.g. the Zero Moment Point \cite{vukobratovic2004zero}.

While walking, references may change and it is not desirable to wait the end of the planned trajectories before changing them. Thus the idea is to ``merge'' two trajectories instead of serializing them. When generating a new trajectory, the robot is supposed to be standing on two feet and the first \textit{switch} needs to last half of its normal time. In view of these considerations, the middle instant of the double support phase is a particularly suitable point to merge two trajectories. The choice of this point eases the merging process since the feet are not supposed to move in that instant, hence the initialization of these trajectories does not need to know their initial desired velocity and/or acceleration. Notice that there may be different \textit{merge points} along the trajectories, depending on the number of (full) double support phases. Two trivial \textit{merge points} are the very first instant of the trajectories themselves and the final point (serialization case).

This method can be assimilated to the \emph{Receding Horizon Principle} \cite{michalska1989receding}. In fact, we plan trajectories for an horizon $T$ but only a portion of them will be actually used, namely the first generated step. This simple strategy allows us to change the unicycle reference trajectory on-line (through the use of a joystick for example) directly affecting the robot motion. In addition, we could correct the new trajectories with the \emph{actual} position of the feet. This is particularly suitable when the foot placement is not perfect, as in torque-controlled walking.

\subsection{The Receding Horizon Controller}
The receding horizon controller used in our architecture inherits from the basic formulation described in Sec. \ref{sec:zmppreview}. Nevertheless, differently from \cite{Kajita2003}, we have added time-varying constraints to bound the ZMP inside the convex hull given by the supporting foot or feet. In this way we increase the robustness of the controller, avoiding overshoots that may cause the robot to tip around one of the foot edges. 
These constraints are modeled as linear inequalities acting on the state variables, i.e.
\begin{align}
\label{eq:constraints_linear}
Z(t) \chi - z(t) \leq 0.
\end{align}
Note that the constraint matrix and vector do depend on the time. In particular, we can exploit the knowledge on the desired feet positions to constraint the ZMP throughout the \emph{whole} horizon, while retaining linearity

To summarize, the full optimal control problem can be represented by the following minimisation problem:
\begin{equation}
\label{eq:mpc_linear}
\begin{aligned}
\min_{\chi ,u } & \int_0^{t_f} \norm{x_\text{ZMP}^*- C_{\text{ZMP}} \chi}_Q^2  + \norm{u}^2_R \dif \tau . \\
\text{s.t.} &~ \dot{\chi} = A \chi + B u, \quad\forall t \in [t_0, t_f) \\
&~ Z(t) \chi - z(t) \leq 0\\
&~ \chi(0) = \bar{\chi}.
\end{aligned}
\end{equation}
The problem in Eq. \eqref{eq:mpc_linear} is solved at each control sampling time by means of a Direct Multiple Shooting method \cite{Bock84,diehl2006fast}. This choice has been driven by the necessity of formulating state constraints, Eq. \eqref{eq:constraints_linear}, throughout the whole horizon.

Applying the Receding Horizon Principle, only the first output of the MPC controller is used. Since the control input is the CoM jerk, we integrate it in order to obtain a desired CoM velocity and position. The latter is also sent to the inverse kinematics (IK) library \cite[InverseKinematics sub-library]{libiDynTree}, together with the desired feet positions to retrieve the desired joints coordinates. Both the MPC an the IK modules are implemented using IPOpt \cite{IPOpt2006}.

\subsection{The Stack of Tasks Balancing Controller}
\label{sec:sot}
The balancing controller has as objective the stabilization of the center of mass position, the left and right feet positions and orientations by defining joint torques. The velocities associated to these tasks are stacked together into $\Upsilon$:
\begin{equation}
\label{eq:tasks}
\Upsilon = \begin{bmatrix}
\dot{p}_{G}^\top & \text{v}_L^\top & \text{v}_R^\top
\end{bmatrix}^\top
\end{equation}
where $\dot{p}_{G} \in \mathbb{R}^3$ is the linear velocity of the center of mass, frame, $\text{v}_L \in \mathbb{R}^6$ and $\text{v}_R \in \mathbb{R}^6$ are vectors of linear and angular velocities of the frames attached to the left and right feet.

Letting $J_G$, $J_L$ and $J_R$ denote respectively the Jacobians of the center of mass position, left and right foot configurations, $J_{\Upsilon}$ can be defined as a stack of the Jacobians associated to each task:
\begin{equation}
J_{\Upsilon} = 
\begin{bmatrix}
J_{G}^\top & J_{L}^\top & J_{R}^\top
\end{bmatrix}^\top
\end{equation}

Furthermore, the task velocities $\Upsilon$ can be computed from $\nu$ using $\Upsilon = J_{\Upsilon} \nu$. By deriving this expression, the task acceleration is
\begin{equation}
\label{eq:task_acceleration}
\dot{\Upsilon} = \dot{J}_{\Upsilon} \nu + J_{\Upsilon} \dot{\nu}
\end{equation}

In view of~\eqref{eq:system} and~\eqref{eq:task_acceleration}, the task accelerations $\dot{\Upsilon}$ can be formulated as a function of the control input $u$ composed of joint torques $\tau$ and contact wrenches $f_k$:
\begin{equation}
\label{eq:task_acceleration_from_u}
\dot{\Upsilon}(u) = \dot{J}_{\Upsilon} \nu + J_{\Upsilon} M^{-1} (B \tau + \sum_{k = 1}^{n_c} {J}^\top_{\mathcal{C}_k} f_k - h).
\end{equation}
In our formulation we want $\dot{\Upsilon}(u)$ to track a specified task acceleration $\dot{\Upsilon}^*$, namely:
\begin{equation}
\label{eq:task_constraint}
	\dot{\Upsilon}(u) = \dot{\Upsilon}^*.
\end{equation} 
The reference accelerations $\dot{\Upsilon}^*$ are computed using a simple PD control strategy for what concerns the linear terms. In parallel, the rotational terms are obtained by adopting a PD controller on $SO(3)$ (see \cite[Section 5.11.6, p.173]{olfati2001nonlinear}), thus avoiding to use a parametrization like Euler angles.

While the task defined above can be considered as \emph{high priority} tasks, we may conceive other possible objectives to shape the robot motion. One example regards the torso orientation, which we would like to keep in an upright posture. Similarly to what have been done above, we have $\mathcal{T} = J_\mathcal{T}\nu$ with $J_\mathcal{T}$ the torso Jacobian. By differentiation we would obtain a result similar to Eq. \eqref{eq:task_acceleration_from_u}. Since this task is considered as \emph{low priority}, we are interested in minimizing the squared Euclidean distance of $\dot{\mathcal{T}}(u)$ from a desired value $\dot{\mathcal{T}}^*$, i.e
\begin{equation}
\label{eq:torso_task}
	\min_{\tau,f_k} \frac{1}{2}\|\dot{\mathcal{T}}(u) - \dot{\mathcal{T}}^*\|^2.
\end{equation}
The reference acceleration $\dot{\mathcal{T}}^*$ is obtained through a PD controller on $SO(3)$.

Finally, we also added another lower priority \emph{postural} task to track joint configurations. Similarly as before, the joint accelerations $\ddot{s}(u)$, which depends upon the control inputs through the dynamics equations Eq. \eqref{eq:system}, are kept close to a reference $\ddot{s}^*$:
\begin{equation}
\label{eq:postural}
\min_{\tau,f_k} \frac{1}{2}\|\ddot{s}(u) - \ddot{s}^*\|^2.
\end{equation}
The postural desired accelerations are defined through a \textit{PD+gravity compensation} control law, \cite{Nava2016}.

Considering the contact wrenches $f_k$ as control inputs, we need to ensure their actual feasibility given the contacts. They are modeled as unilateral, with a limited friction coefficient. An additional condition regards the Center of Pressure, CoP, which needs to lie inside the contact patch. All these conditions can be grouped in a set of inequality constraints applied to the contact wrenches
\begin{equation}
\label{eq:contact_constraint}
	C f_k \leq b \quad \forall k \leq n_c.
\end{equation}
Finally, we can group Eq.s \eqref{eq:task_constraint}$-$\eqref{eq:contact_constraint} in the following QP formulation:
\begin{IEEEeqnarray}{RRCL}
	\IEEEyesnumber \phantomsection	\label{eq:qp_balancing}
\min{\tau,f_k} & & \Gamma&\IEEEyessubnumber\\
	\text{s.t.}: & \dot{\Upsilon}(u) &=& \dot{\Upsilon}^* \IEEEyessubnumber \\
		&C f_k &\leq& b \quad \forall k \leq n_c \IEEEyessubnumber
\end{IEEEeqnarray} 
where
$
\Gamma = \frac{1}{2}\|\ddot{s}(u) - \ddot{s}^*\|^2 +  \frac{w_\mathcal{T}}{2}\|\dot{\mathcal{T}}(u) - \dot{\mathcal{T}}^*\|^2 +  \frac{w_\tau}{2}\|\tau\|^2.
$
With respect to Eq. \eqref{eq:torso_task} and \eqref{eq:postural}, an additional term is inserted, namely the 2-norm of the joint torques, which serves as a regularization. Among several feasible solutions, we are mostly interested in the one which requires the least effort. By changing the weights $w$ we can tune the relative importance of each term.

Let us focus on the feet contact wrenches $f_l$ and $f_r$. During the \textit{switch} phases we expect one of the two wrenches to vanish in order to smoothly deactivate the corresponding contact. In order to ease this process we added a cost term in $\Gamma$, equal to
$
	\frac{w_f}{2}\left(\mathcal{F}_r\|f_l\| + \mathcal{F}_l\|f_r\| \right)
$
where $\mathcal{F}_r$ (respectively $\mathcal{F}_l$) is the normalized load that the right (respectively left) foot is supposed to carry. It is $1$ when the corresponding foot is supposed to hold the full weight of the robot, $0$ when unloaded. This information is provided by the planner described in Sec \ref{sec:footstepPlanner}. 
The QP problem of Eq. \eqref{eq:qp_balancing} is solved at a rate of 100Hz through qpOASES~\cite{Ferreau2014}.

%% file: tex/Experiments.tex
\section{Experiments}
The presented architecture is composed by three different parts. In this section, we are going to show three different experiments whose aim is to test each part in an incremental way. All the experiments have been performed on the iCub robot, controlling 23 Dofs either in position or in torque control. The complete experiments videos are available as multimedia attachements to this paper
\subsection{Test of the RH controller in position control}

\begin{figure}[tpb]
	\centering
	\def\svgwidth{\columnwidth}
	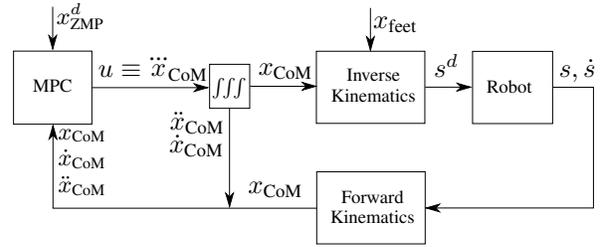
	\caption{A first skeleton of the architecture, composed only by the MPC (RH) controller connected to the inverse kinematics (IK). Their output is directly sent to the robot.} 
	\label{fig:mpc}
\end{figure}

\begin{figure}[tpb]
	\centering
	\subfloat[$t=t_0$] {\includegraphics[width=.22\columnwidth]{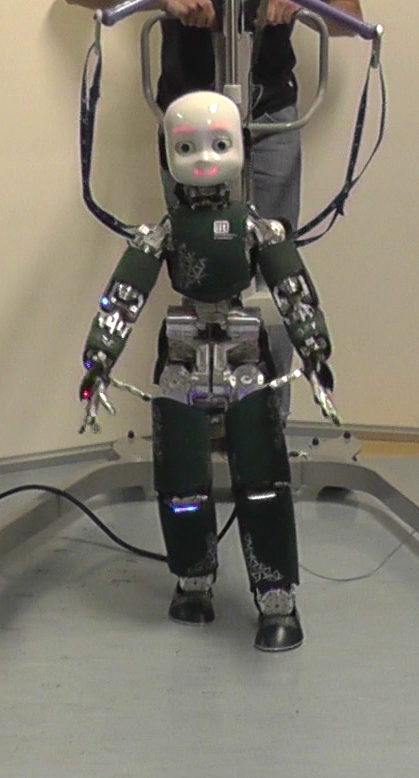}}
	\subfloat[$t=t_0+1s$] {\includegraphics[width=.22\columnwidth]{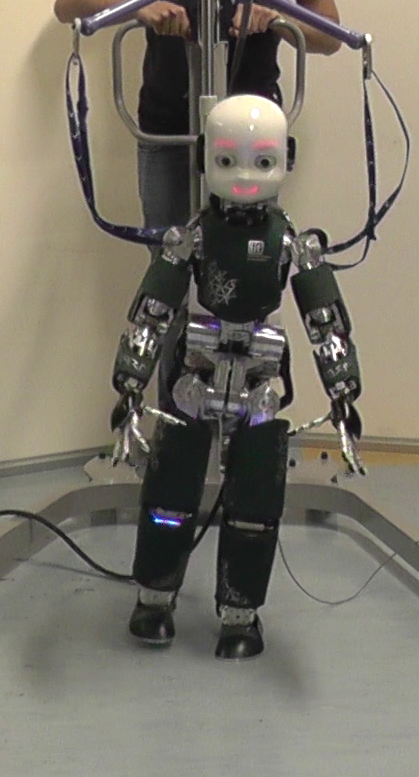}}
	\subfloat[$t=t_0+2s$] {\includegraphics[width=.22\columnwidth]{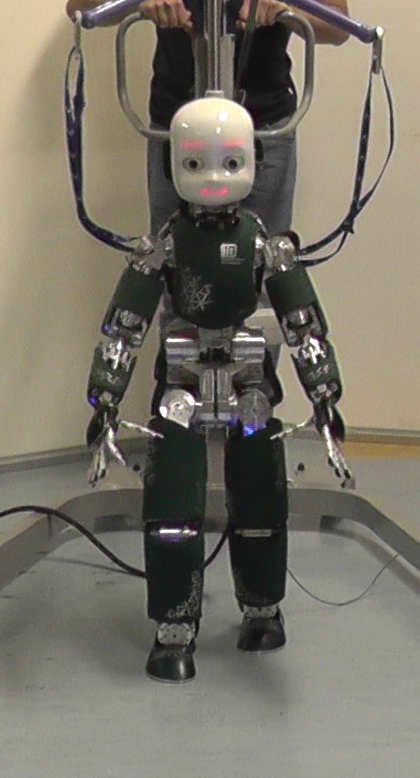}}
	\subfloat[$t=t_0+3s$] {\includegraphics[width=.22\columnwidth]{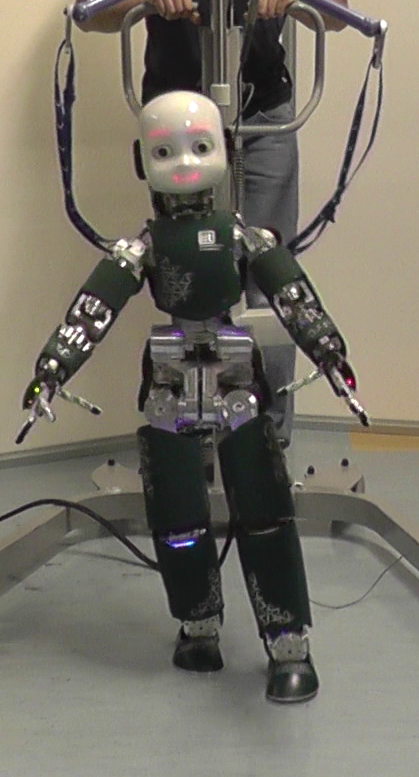}}
	\caption{Snapshots extracted from the accompanying video. The robot is walking straight in position control.}
	\label{fig:exp_straight}
\end{figure}

In the first experiment we used the RH module to control the robot joints in \emph{position control}. As depicted in the architecture of Figure \ref{fig:mpc} the control loop is closed on the CoM position only, avoiding to stream in the controller noisy measurements like joints velocities. The desired feet trajectories $x_\text{feet}$ and the desired ZMP profile $x_{\text{ZMP}}^d$ are obtained by an offline planner described in \cite{hu2016walking}. 

Both the RHC and the IK are running on the iCub head, which is equipped with a 4$^{th}$ generation Intel\textsuperscript{\textregistered} Core i7@1.7GHz and 8GB of RAM. The whole architecture takes in average less than 8$\mathrm{ms}$ for a control loop.

The robot is taking steps of $14\mathrm{cm}$ long in $1.25s$ (of which $0.65s$ in double support). Fig. \ref{fig:exp_straight} presents some snapshots of the robot while walking.

\subsection{Adding the unicycle planner}
\begin{figure}[tpb]
	\centering
	\def\svgwidth{\columnwidth}
	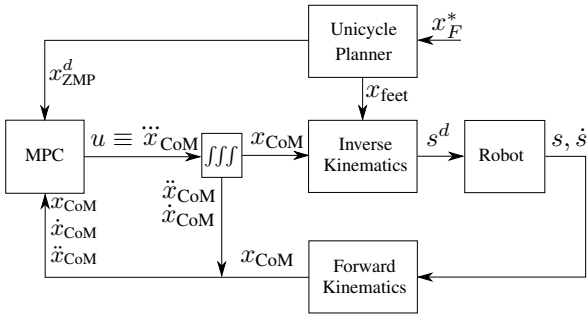
	\caption{The scheme of Figure \ref{fig:mpc} has been upgrade with the unicycle planner which is responsible of providing online references.}
	\label{fig:mpc+unicycle}
\end{figure}
\begin{figure}[tpb]
	\centering
	\subfloat[$t=t_0$] {\includegraphics[width=.22\columnwidth]{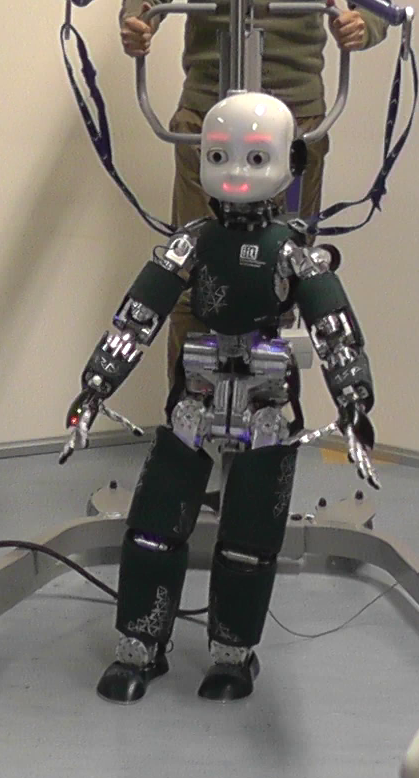}}
	\subfloat[$t=t_0+1s$] {\includegraphics[width=.22\columnwidth]{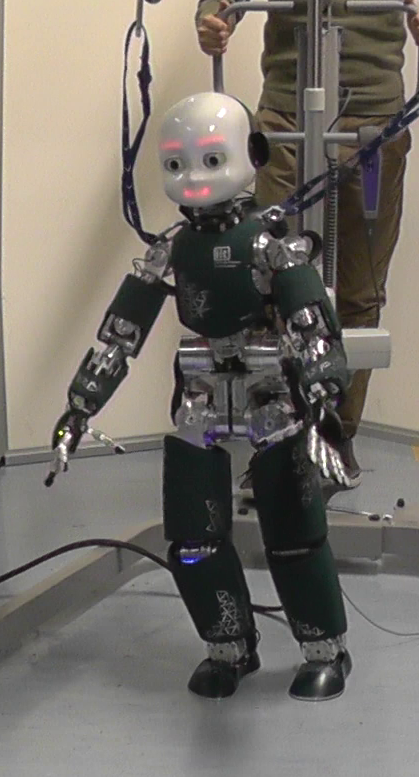}}
	\subfloat[$t=t_0+2s$] {\includegraphics[width=.22\columnwidth]{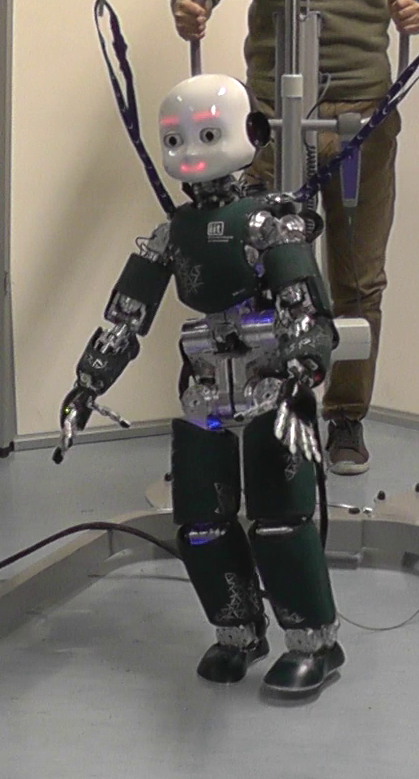}}
	\subfloat[$t=t_0+3s$] {\includegraphics[width=.22\columnwidth]{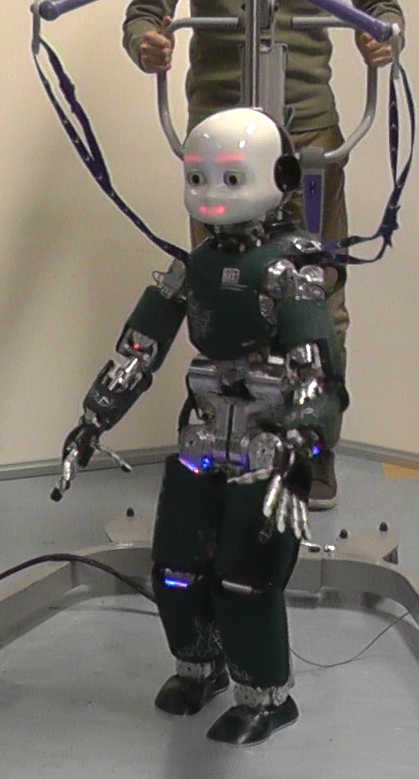}}
	\caption{Thanks to the unicycle planner, the robot walks while turning right.}
	\label{fig:exp_unicycle}
\end{figure}

The controller has been improved by connecting the unicycle planner, as shown in Fig. \ref{fig:mpc+unicycle}. This allow us to adapt the robot walking direction in an \emph{online} fashion. As an example, by using a joystick we can drive a reference position $x^*_F$ for the point $F$ attached to the unicycle. Depending on how much we press the joypad, this point moves further away from the robot.

Using this planner, the step timings and dimensions are not fixed a priori. In this particular experiment, a single step could last between $1.3s$ and $5.0s$, while the swing foot can land in a radius of $0.175m$ from the stance foot. Fig. \ref{fig:exp_unicycle} presents some snapshots of the robot while negotiating a right turn.

\subsection{Complete Architecture}

\begin{figure}[tpb]
	\centering
	\def\svgwidth{\columnwidth}
	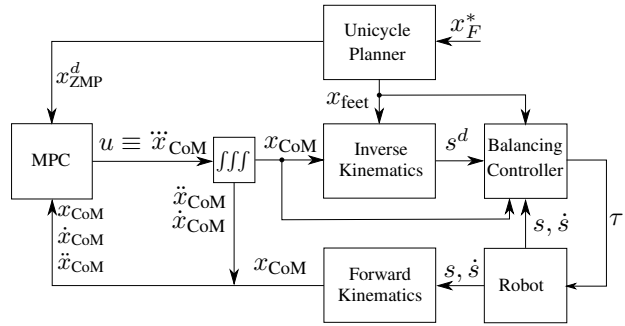
	\caption{The complete architecture. The output of the IK is no longer sent to the robot, but to the stack-of-task balancing controller, together with the desired CoM and feet position. Joint positions and velocities are taken as feedback from the robot.}
	\label{fig:complete}
\end{figure}

\begin{figure}[tpb]
	\centering
	\includegraphics[width=0.9\columnwidth]{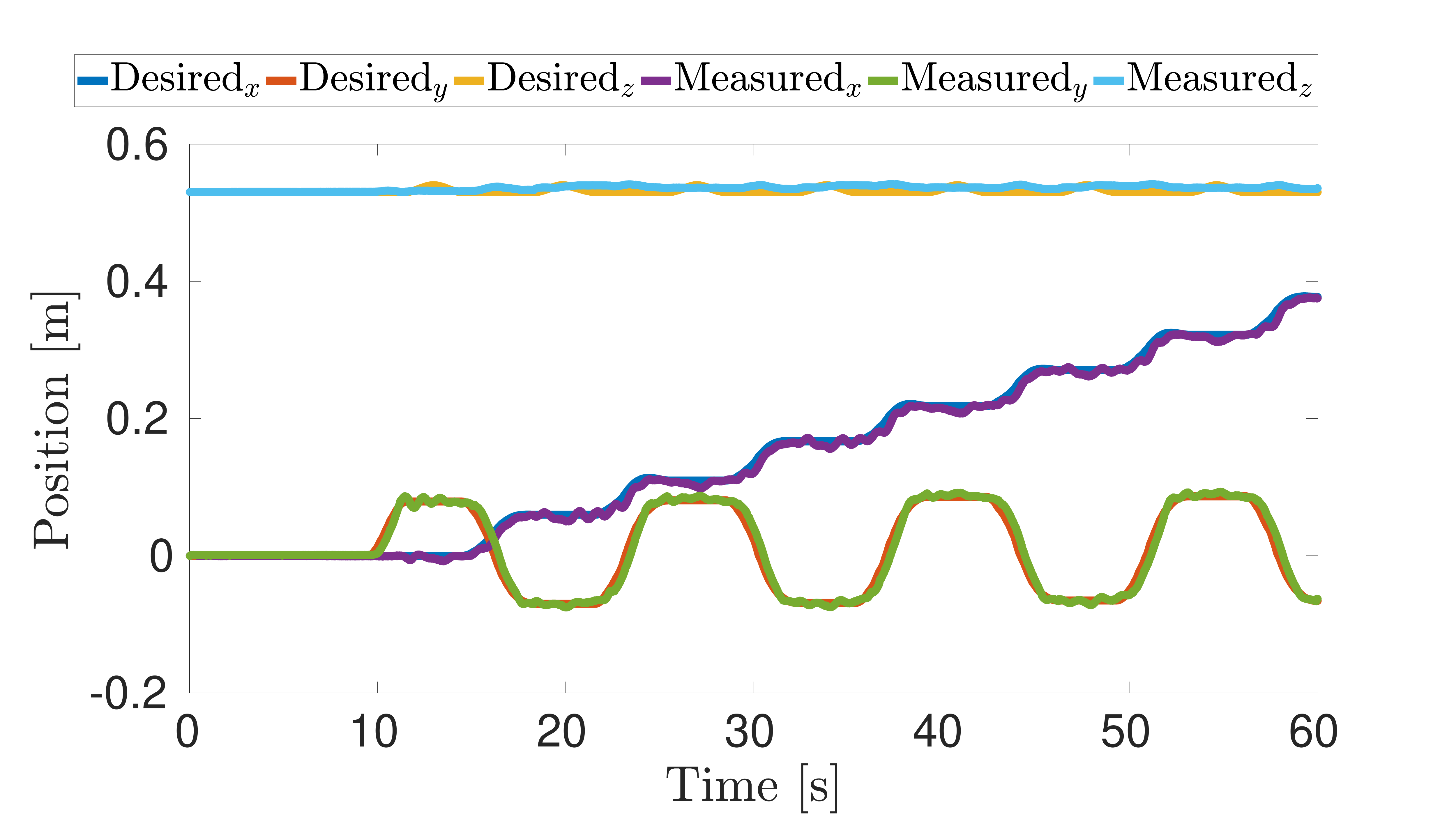}
	\caption{Center of Mass (CoM) tracking when taking some steps in torque control. The quantities are expressed in an inertial frame $\mathcal{I}$}
	\label{fig:com}
\end{figure}

\begin{figure}[tpb]
	\centering
	\includegraphics[width=0.9\columnwidth]{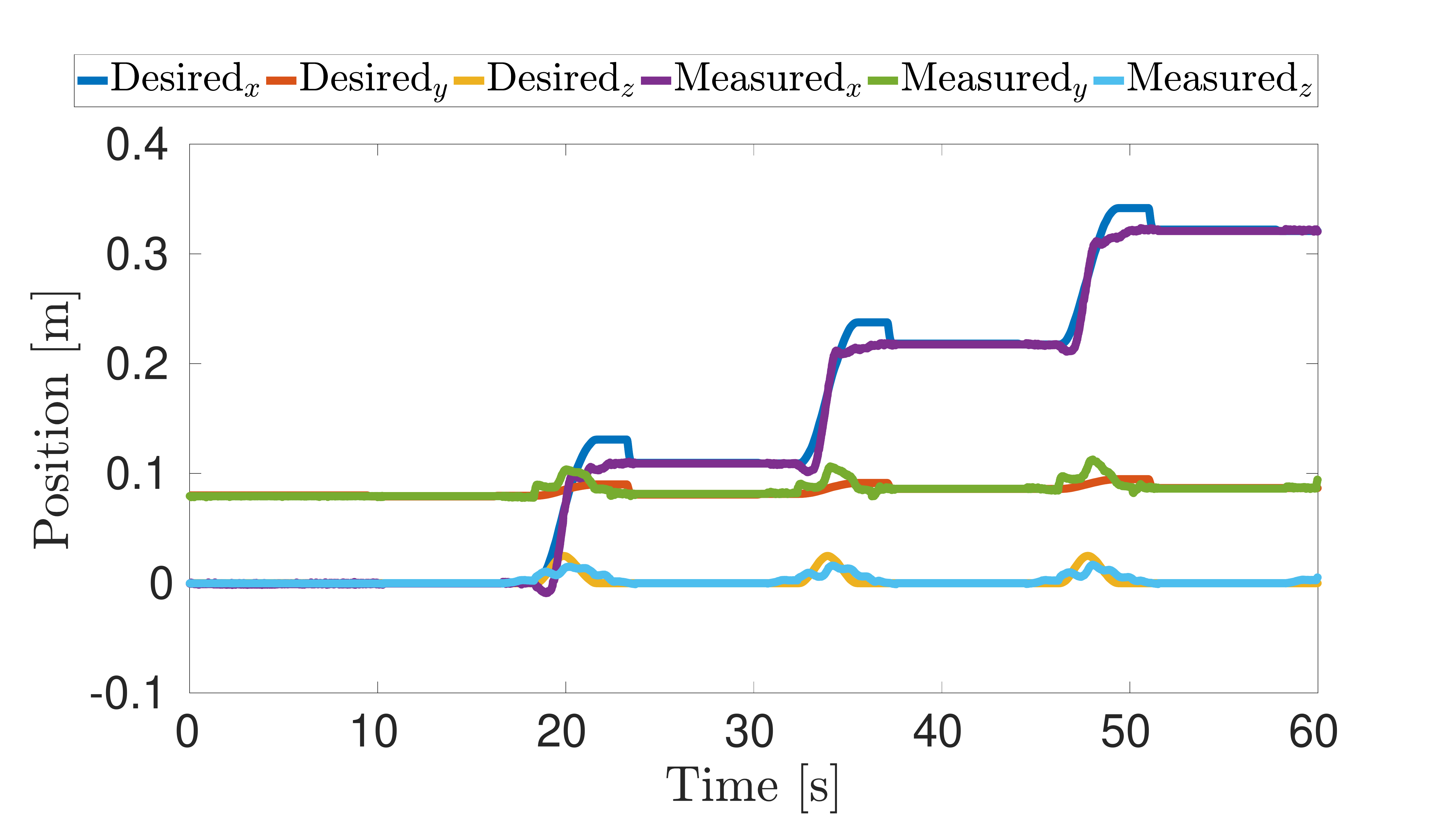}
	\caption{Tracking of the left foot positions. Notice that after the step is taken, the \emph{desired} values are updated according to the measured position.}
	\label{fig:lFoot}
\end{figure}

Finally, we plugged also the task based balancing controller presented in Sec. \ref{sec:sot}. The overall architecture is depicted in Figure \ref{fig:complete}, highlighting that the stack of task balancing controller is now in charge of sending commands the robot. In order to draw comparisons with the first experiments, we followed again a straight trajectory. Even if the task is similar, it is necessary to use the Unicycle Planner now in order to cope with feet positioning errors (see Fig. \ref{fig:lFoot}). In fact, trajectories can be updated in order to take into account possible feet misplacements. Fig. \ref{fig:com} shows the CoM tracking capabilities of the presented balancing controller. During this experiment, the minimum stepping time had to be increased to $3s$ (the maximum is still $5s$), while maintaining the same maximum step length of the previous experiment. Even if we had to slow down a bit the walking speed, the results are still promising. While walking, the robot is much smoother in its motion, reducing the probabilities of falling in case of an unforeseen disturbances.

%% file: figures/walking_schema.pdf_tex
\begingroup%
  \makeatletter%
  \providecommand\color[2][]{%
    \errmessage{(Inkscape) Color is used for the text in Inkscape, but the package 'color.sty' is not loaded}%
    \renewcommand\color[2][]{}%
  }%
  \providecommand\transparent[1]{%
    \errmessage{(Inkscape) Transparency is used (non-zero) for the text in Inkscape, but the package 'transparent.sty' is not loaded}%
    \renewcommand\transparent[1]{}%
  }%
  \providecommand\rotatebox[2]{#2}%
  \ifx\svgwidth\undefined%
    \setlength{\unitlength}{547.17773201bp}%
    \ifx\svgscale\undefined%
      \relax%
    \else%
      \setlength{\unitlength}{\unitlength * \real{\svgscale}}%
    \fi%
  \else%
    \setlength{\unitlength}{\svgwidth}%
  \fi%
  \global\let\svgwidth\undefined%
  \global\let\svgscale\undefined%
  \makeatother%
  \begin{picture}(1,0.36601187)%
    \put(0,0){\includegraphics[width=\unitlength,page=1]{walking_schema.pdf}}%
    \put(0.03058775,0.23439433){\color[rgb]{0,0,0}\makebox(0,0)[lb]{\smash{\scriptsize MPC}}}%
    \put(0.30538505,0.23432421){\color[rgb]{0,0,0}\makebox(0,0)[lb]{\smash{\scriptsize$\iiint$}}}%
    \put(0,0){\includegraphics[width=\unitlength,page=2]{walking_schema.pdf}}%
    \put(0.48984803,0.2192155){\color[rgb]{0,0,0}\makebox(0,0)[lb]{\smash{\scriptsize\shortstack[c]{Inverse\\Kinematics}}}}%
    \put(0,0){\includegraphics[width=\unitlength,page=3]{walking_schema.pdf}}%
    \put(0.72892452,0.23413325){\color[rgb]{0,0,0}\makebox(0,0)[lb]{\smash{\scriptsize Robot}}}%
    \put(0.31067436,0.24379321){\color[rgb]{0,0,0}\makebox(0,0)[lb]{\smash{}}}%
    \put(0,0){\includegraphics[width=\unitlength,page=4]{walking_schema.pdf}}%
    \put(0.13107993,0.2583689){\color[rgb]{0,0,0}\makebox(0,0)[lb]{\smash{$u\equiv\dddot{x}_\text{CoM}$}}}%
    \put(-0.03865066,0.25632505){\color[rgb]{0,0,0}\makebox(0,0)[lb]{\smash{}}}%
    \put(0,0){\includegraphics[width=\unitlength,page=5]{walking_schema.pdf}}%
    \put(0.24171512,0.17724128){\color[rgb]{0,0,0}\makebox(0,0)[lb]{\smash{$\ddot{x}_\text{CoM}$}}}%
    \put(0.24038746,0.14116616){\color[rgb]{0,0,0}\makebox(0,0)[lb]{\smash{$\dot{x}_\text{CoM}$}}}%
    \put(0.37761525,0.26112893){\color[rgb]{0,0,0}\makebox(0,0)[lb]{\smash{${x}_\text{CoM}$}}}%
    \put(0,0){\includegraphics[width=\unitlength,page=6]{walking_schema.pdf}}%
    \put(0.06697493,0.33869832){\color[rgb]{0,0,0}\makebox(0,0)[lb]{\smash{\small$x_\text{ZMP}^d$}}}%
    \put(0.55467598,0.33384832){\color[rgb]{0,0,0}\makebox(0,0)[lb]{\smash{$x_{\text{feet}}$}}}%
    \put(0.64788519,0.25815263){\color[rgb]{0,0,0}\makebox(0,0)[lb]{\smash{$s^d$}}}%
    \put(0.84034524,0.25789156){\color[rgb]{0,0,0}\makebox(0,0)[lb]{\smash{$s, \dot{s}$}}}%
    \put(0,0){\includegraphics[width=\unitlength,page=7]{walking_schema.pdf}}%
    \put(0.48836537,0.032214){\color[rgb]{0,0,0}\makebox(0,0)[lb]{\smash{\scriptsize\shortstack[c]{Forward\\Kinematics}}}}%
    \put(0.06935488,0.07988243){\color[rgb]{0,0,0}\makebox(0,0)[lb]{\smash{\small\shortstack[c]{${x}_\text{CoM}$\\$\dot{x}_\text{CoM}$\\$\ddot{x}_\text{CoM}$}}}}%
    \put(0,0){\includegraphics[width=\unitlength,page=8]{walking_schema.pdf}}%
    \put(0.36204297,0.07888767){\color[rgb]{0,0,0}\makebox(0,0)[lb]{\smash{${x}_\text{CoM}$}}}%
    \put(0,0){\includegraphics[width=\unitlength,page=9]{walking_schema.pdf}}%
  \end{picture}%
\endgroup%

%% file: figures/walking_schema_unicycle.pdf_tex
\begingroup%
  \makeatletter%
  \providecommand\color[2][]{%
    \errmessage{(Inkscape) Color is used for the text in Inkscape, but the package 'color.sty' is not loaded}%
    \renewcommand\color[2][]{}%
  }%
  \providecommand\transparent[1]{%
    \errmessage{(Inkscape) Transparency is used (non-zero) for the text in Inkscape, but the package 'transparent.sty' is not loaded}%
    \renewcommand\transparent[1]{}%
  }%
  \providecommand\rotatebox[2]{#2}%
  \ifx\svgwidth\undefined%
    \setlength{\unitlength}{547.17773201bp}%
    \ifx\svgscale\undefined%
      \relax%
    \else%
      \setlength{\unitlength}{\unitlength * \real{\svgscale}}%
    \fi%
  \else%
    \setlength{\unitlength}{\svgwidth}%
  \fi%
  \global\let\svgwidth\undefined%
  \global\let\svgscale\undefined%
  \makeatother%
  \begin{picture}(1,0.47552452)%
    \put(0,0){\includegraphics[width=\unitlength,page=1]{walking_schema_unicycle.pdf}}%
    \put(0.03058775,0.23439433){\color[rgb]{0,0,0}\makebox(0,0)[lb]{\smash{\scriptsize MPC}}}%
    \put(0.30538505,0.23432421){\color[rgb]{0,0,0}\makebox(0,0)[lb]{\smash{\scriptsize$\iiint$}}}%
    \put(0,0){\includegraphics[width=\unitlength,page=2]{walking_schema_unicycle.pdf}}%
    \put(0.48984803,0.2192155){\color[rgb]{0,0,0}\makebox(0,0)[lb]{\smash{\scriptsize\shortstack[c]{Inverse\\Kinematics}}}}%
    \put(0,0){\includegraphics[width=\unitlength,page=3]{walking_schema_unicycle.pdf}}%
    \put(0.72892452,0.23413325){\color[rgb]{0,0,0}\makebox(0,0)[lb]{\smash{\scriptsize Robot}}}%
    \put(0.31067436,0.24379321){\color[rgb]{0,0,0}\makebox(0,0)[lb]{\smash{}}}%
    \put(0,0){\includegraphics[width=\unitlength,page=4]{walking_schema_unicycle.pdf}}%
    \put(0.13107993,0.2583689){\color[rgb]{0,0,0}\makebox(0,0)[lb]{\smash{$u\equiv\dddot{x}_\text{CoM}$}}}%
    \put(-0.03865066,0.25632505){\color[rgb]{0,0,0}\makebox(0,0)[lb]{\smash{}}}%
    \put(0,0){\includegraphics[width=\unitlength,page=5]{walking_schema_unicycle.pdf}}%
    \put(0.24171512,0.17724128){\color[rgb]{0,0,0}\makebox(0,0)[lb]{\smash{$\ddot{x}_\text{CoM}$}}}%
    \put(0.24038746,0.14116616){\color[rgb]{0,0,0}\makebox(0,0)[lb]{\smash{$\dot{x}_\text{CoM}$}}}%
    \put(0.37761525,0.26112893){\color[rgb]{0,0,0}\makebox(0,0)[lb]{\smash{${x}_\text{CoM}$}}}%
    \put(0,0){\includegraphics[width=\unitlength,page=6]{walking_schema_unicycle.pdf}}%
    \put(0.06697493,0.35916699){\color[rgb]{0,0,0}\makebox(0,0)[lb]{\smash{\small$x_\text{ZMP}^d$}}}%
    \put(0.55467598,0.33384832){\color[rgb]{0,0,0}\makebox(0,0)[lb]{\smash{$x_{\text{feet}}$}}}%
    \put(0.64788519,0.25815263){\color[rgb]{0,0,0}\makebox(0,0)[lb]{\smash{$s^d$}}}%
    \put(0.84034524,0.25789156){\color[rgb]{0,0,0}\makebox(0,0)[lb]{\smash{$s, \dot{s}$}}}%
    \put(0,0){\includegraphics[width=\unitlength,page=7]{walking_schema_unicycle.pdf}}%
    \put(0.48836537,0.032214){\color[rgb]{0,0,0}\makebox(0,0)[lb]{\smash{\scriptsize\shortstack[c]{Forward\\Kinematics}}}}%
    \put(0.06935488,0.07988243){\color[rgb]{0,0,0}\makebox(0,0)[lb]{\smash{\small\shortstack[c]{${x}_\text{CoM}$\\$\dot{x}_\text{CoM}$\\$\ddot{x}_\text{CoM}$}}}}%
    \put(0,0){\includegraphics[width=\unitlength,page=8]{walking_schema_unicycle.pdf}}%
    \put(0.36204297,0.07888767){\color[rgb]{0,0,0}\makebox(0,0)[lb]{\smash{${x}_\text{CoM}$}}}%
    \put(0,0){\includegraphics[width=\unitlength,page=9]{walking_schema_unicycle.pdf}}%
    \put(0.50057431,0.44803592){\color[rgb]{0,0,0}\makebox(0,0)[lt]{\begin{minipage}{0.42700104\unitlength}\raggedright \scriptsize\shortstack[c]{Unicycle\\Planner}\end{minipage}}}%
    \put(0,0){\includegraphics[width=\unitlength,page=10]{walking_schema_unicycle.pdf}}%
    \put(0.65608368,0.43618517){\color[rgb]{0,0,0}\makebox(0,0)[lb]{\smash{$x_{F}^*$}}}%
  \end{picture}%
\endgroup%

%% file: figures/walking_schema_torque.pdf_tex
\begingroup%
  \makeatletter%
  \providecommand\color[2][]{%
    \errmessage{(Inkscape) Color is used for the text in Inkscape, but the package 'color.sty' is not loaded}%
    \renewcommand\color[2][]{}%
  }%
  \providecommand\transparent[1]{%
    \errmessage{(Inkscape) Transparency is used (non-zero) for the text in Inkscape, but the package 'transparent.sty' is not loaded}%
    \renewcommand\transparent[1]{}%
  }%
  \providecommand\rotatebox[2]{#2}%
  \ifx\svgwidth\undefined%
    \setlength{\unitlength}{531.05693123bp}%
    \ifx\svgscale\undefined%
      \relax%
    \else%
      \setlength{\unitlength}{\unitlength * \real{\svgscale}}%
    \fi%
  \else%
    \setlength{\unitlength}{\svgwidth}%
  \fi%
  \global\let\svgwidth\undefined%
  \global\let\svgscale\undefined%
  \makeatother%
  \begin{picture}(1,0.49059625)%
    \put(0,0){\includegraphics[width=\unitlength,page=1]{walking_schema_torque.pdf}}%
    \put(0.03151627,0.2421463){\color[rgb]{0,0,0}\makebox(0,0)[lb]{\smash{\scriptsize MPC}}}%
    \put(0.31465534,0.24207406){\color[rgb]{0,0,0}\makebox(0,0)[lb]{\smash{\scriptsize$\iiint$}}}%
    \put(0,0){\includegraphics[width=\unitlength,page=2]{walking_schema_torque.pdf}}%
    \put(0.50471789,0.2265067){\color[rgb]{0,0,0}\makebox(0,0)[lb]{\smash{\scriptsize\shortstack[c]{Inverse\\Kinematics}}}}%
    \put(0,0){\includegraphics[width=\unitlength,page=3]{walking_schema_torque.pdf}}%
    \put(0.73297465,0.22681299){\color[rgb]{0,0,0}\makebox(0,0)[lb]{\smash{\scriptsize\shortstack[c]{Balancing\\Controller}}}}%
    \put(0.32010521,0.2518305){\color[rgb]{0,0,0}\makebox(0,0)[lb]{\smash{}}}%
    \put(0,0){\includegraphics[width=\unitlength,page=4]{walking_schema_torque.pdf}}%
    \put(0.13505901,0.26684864){\color[rgb]{0,0,0}\makebox(0,0)[lb]{\smash{$u\equiv\dddot{x}_\text{CoM}$}}}%
    \put(-0.03982394,0.26474275){\color[rgb]{0,0,0}\makebox(0,0)[lb]{\smash{}}}%
    \put(0,0){\includegraphics[width=\unitlength,page=5]{walking_schema_torque.pdf}}%
    \put(0.24905264,0.18325831){\color[rgb]{0,0,0}\makebox(0,0)[lb]{\smash{$\ddot{x}_\text{CoM}$}}}%
    \put(0.24768468,0.14608809){\color[rgb]{0,0,0}\makebox(0,0)[lb]{\smash{$\dot{x}_\text{CoM}$}}}%
    \put(0.38907816,0.26969246){\color[rgb]{0,0,0}\makebox(0,0)[lb]{\smash{${x}_\text{CoM}$}}}%
    \put(0,0){\includegraphics[width=\unitlength,page=6]{walking_schema_torque.pdf}}%
    \put(0.06900803,0.37070657){\color[rgb]{0,0,0}\makebox(0,0)[lb]{\smash{\small$x_\text{ZMP}^d$}}}%
    \put(0.48477268,0.33821111){\color[rgb]{0,0,0}\makebox(0,0)[lb]{\smash{$x_{\text{feet}}$}}}%
    \put(0.66755243,0.26662581){\color[rgb]{0,0,0}\makebox(0,0)[lb]{\smash{$s^d$}}}%
    \put(0.6653708,0.0718496){\color[rgb]{0,0,0}\makebox(0,0)[lb]{\smash{$s, \dot{s}$}}}%
    \put(0,0){\includegraphics[width=\unitlength,page=7]{walking_schema_torque.pdf}}%
    \put(0.50319022,0.03382857){\color[rgb]{0,0,0}\makebox(0,0)[lb]{\smash{\scriptsize\shortstack[c]{Forward\\Kinematics}}}}%
    \put(0.07146023,0.08294403){\color[rgb]{0,0,0}\makebox(0,0)[lb]{\smash{\small\shortstack[c]{${x}_\text{CoM}$\\$\dot{x}_\text{CoM}$\\$\ddot{x}_\text{CoM}$}}}}%
    \put(0,0){\includegraphics[width=\unitlength,page=8]{walking_schema_torque.pdf}}%
    \put(0.37303317,0.08191907){\color[rgb]{0,0,0}\makebox(0,0)[lb]{\smash{${x}_\text{CoM}$}}}%
    \put(0,0){\includegraphics[width=\unitlength,page=9]{walking_schema_torque.pdf}}%
    \put(0.51275692,0.46227321){\color[rgb]{0,0,0}\makebox(0,0)[lt]{\begin{minipage}{0.43996311\unitlength}\raggedright \scriptsize\shortstack[c]{Unicycle\\Planner}\end{minipage}}}%
    \put(0,0){\includegraphics[width=\unitlength,page=10]{walking_schema_torque.pdf}}%
    \put(0.67599981,0.45006272){\color[rgb]{0,0,0}\makebox(0,0)[lb]{\smash{$x_{F}^*$}}}%
    \put(0,0){\includegraphics[width=\unitlength,page=11]{walking_schema_torque.pdf}}%
    \put(0.75075255,0.0502502){\color[rgb]{0,0,0}\makebox(0,0)[lb]{\smash{\scriptsize Robot}}}%
    \put(0,0){\includegraphics[width=\unitlength,page=12]{walking_schema_torque.pdf}}%
    \put(0.80339992,0.14232904){\color[rgb]{0,0,0}\makebox(0,0)[lb]{\smash{$s, \dot{s}$}}}%
    \put(0.92052507,0.15176049){\color[rgb]{0,0,0}\makebox(0,0)[lb]{\smash{$\tau$}}}%
    \put(0,0){\includegraphics[width=\unitlength,page=13]{walking_schema_torque.pdf}}%
  \end{picture}%
\endgroup%

%% file: tex/Conclusions.tex
\section{Conclusions}
This architecture is able to cope with variable walking speed and advance while turning while keeping certain degree of compliance to better adapt to the  \emph{online} changes of desired reference trajectories. Focusing on torque control, experiments have shown some bottlenecks that, for the time being, prevented to achieve higher performances. These issue are mainly  related to the estimation of joint torques (limiting the performances of the joint torque controller) and to the floating base estimation. As a future work, we plan to increase the robustness of the architecture against these uncertainties.

Even if the results presented on this paper have to be considered as preliminary, they enlighten the flexibility properties of the proposed architecture. The result on torque control walking is promising, since its inherent compliance can increase the robustness of the walking motion, against, for example, un-modeled ground slope variations.

%% file: root.bbl
\begin{thebibliography}{10}
\providecommand{\url}[1]{#1}
\csname url@rmstyle\endcsname
\providecommand{\newblock}{\relax}
\providecommand{\bibinfo}[2]{#2}
\providecommand\BIBentrySTDinterwordspacing{\spaceskip=0pt\relax}
\providecommand\BIBentryALTinterwordstretchfactor{4}
\providecommand\BIBentryALTinterwordspacing{\spaceskip=\fontdimen2\font plus
\BIBentryALTinterwordstretchfactor\fontdimen3\font minus
  \fontdimen4\font\relax}
\providecommand\BIBforeignlanguage[2]{{%
\expandafter\ifx\csname l@#1\endcsname\relax
\typeout{** WARNING: IEEEtran.bst: No hyphenation pattern has been}%
\typeout{** loaded for the language `#1'. Using the pattern for}%
\typeout{** the default language instead.}%
\else
\language=\csname l@#1\endcsname
\fi
#2}}

\bibitem{feng2015optimization}
S.~Feng, E.~Whitman, X.~Xinjilefu, and C.~G. Atkeson, ``Optimization-based full
  body control for the darpa robotics challenge,'' \emph{Journal of Field
  Robotics}, vol.~32, no.~2, pp. 293--312, 2015.

\bibitem{dai2014whole}
H.~Dai, A.~Valenzuela, and R.~Tedrake, ``Whole-body motion planning with
  centroidal dynamics and full kinematics,'' in \emph{2014 IEEE-RAS
  International Conference on Humanoid Robots}, pp. 295--302.

\bibitem{herzog2015trajectory}
A.~Herzog, N.~Rotella, S.~Schaal, and L.~Righetti, ``Trajectory generation for
  multi-contact momentum control,'' in \emph{Humanoid Robots (Humanoids), 2015
  IEEE-RAS 15th International Conference on}.\hskip 1em plus 0.5em minus
  0.4em\relax IEEE, 2015, pp. 874--880.

\bibitem{carpentier2016versatile}
J.~Carpentier, S.~Tonneau, M.~Naveau, O.~Stasse, and N.~Mansard, ``A versatile
  and efficient pattern generator for generalized legged locomotion,'' in
  \emph{Robotics and Automation (ICRA), 2016 IEEE International Conference
  on}.\hskip 1em plus 0.5em minus 0.4em\relax IEEE, 2016, pp. 3555--3561.

\bibitem{deits2014footstep}
R.~Deits and R.~Tedrake, ``Footstep planning on uneven terrain with
  mixed-integer convex optimization,'' in \emph{Humanoid Robots (Humanoids),
  2014 14th IEEE-RAS International Conference on}.

\bibitem{flavigne2010reactive}
D.~Flavigne, J.~Pettr{\'e}e, K.~Mombaur, J.-P. Laumond, \emph{et~al.},
  ``Reactive synthesizing of human locomotion combining nonholonomic and
  holonomic behaviors,'' in \emph{Biomedical Robotics and Biomechatronics
  (BioRob), 2010 3rd IEEE RAS and EMBS International Conference on}.\hskip 1em
  plus 0.5em minus 0.4em\relax IEEE, 2010, pp. 632--637.

\bibitem{mombaur2010human}
K.~Mombaur, A.~Truong, and J.-P. Laumond, ``From human to humanoid
  locomotion—an inverse optimal control approach,'' \emph{Autonomous robots},
  vol.~28, no.~3, pp. 369--383, 2010.

\bibitem{handford2014sideways}
M.~L. Handford and M.~Srinivasan, ``Sideways walking: preferred is slow, slow
  is optimal, and optimal is expensive,'' \emph{Biology letters}, vol.~10,
  no.~1, p. 20131006, 2014.

\bibitem{PascalHandbook}
P.~Morin and C.~Samson, \emph{Handbook of Robotics}.\hskip 1em plus 0.5em minus
  0.4em\relax Springer, 2008, ch. Motion control of wheeled mobile robots, pp.
  799--826.

\bibitem{faragasso2013vision}
A.~Faragasso, G.~Oriolo, A.~Paolillo, and M.~Vendittelli, ``Vision-based
  corridor navigation for humanoid robots,'' in \emph{Robotics and Automation
  (ICRA), 2013 IEEE International Conference on}, pp. 3190--3195.

\bibitem{cognetti2016real}
M.~Cognetti, D.~De~Simone, L.~Lanari, and G.~Oriolo, ``Real-time planning and
  execution of evasive motions for a humanoid robot,'' in \emph{Robotics and
  Automation (ICRA), 2016 IEEE International Conference on}.\hskip 1em plus
  0.5em minus 0.4em\relax IEEE, 2016, pp. 4200--4206.

\bibitem{Mayne2000Stability}
D.~Mayne, J.~Rawlings, C.~Rao, and P.~Scokaert, ``Constrained model predictive
  control: Stability and optimality,'' \emph{Automatica}, vol.~36, no.~6, pp.
  789 -- 814, 2000.

\bibitem{Kajita2001}
S.~Kajita, F.~Kanehiro, K.~Kaneko, K.~Yokoi, and H.~Hirukawa, ``The 3d linear
  inverted pendulum mode: a simple modeling for a biped walking pattern
  generation,'' in \emph{Intelligent Robots and Systems, 2001. Proceedings.
  IEEE/RSJ International Conference on}, pp. 239--246.

\bibitem{Pratt2006}
J.~Pratt, J.~Carff, S.~Drakunov, and A.~Goswami, ``Capture point: A step toward
  humanoid push recovery,'' in \emph{Humanoid Robots, 2006 6th IEEE-RAS
  International Conference on}, Dec 2006, pp. 200--207.

\bibitem{diedam2008online}
H.~Diedam, D.~Dimitrov, P.-B. Wieber, K.~Mombaur, and M.~Diehl, ``Online
  walking gait generation with adaptive foot positioning through linear model
  predictive control,'' in \emph{2008 IEEE/RSJ International Conference on
  Intelligent Robots and Systems}.\hskip 1em plus 0.5em minus 0.4em\relax IEEE,
  pp. 1121--1126.

\bibitem{missura2014balanced}
M.~Missura and S.~Behnke, ``Balanced walking with capture steps,'' in
  \emph{Robot Soccer World Cup}.\hskip 1em plus 0.5em minus 0.4em\relax
  Springer, 2014, pp. 3--15.

\bibitem{bombile2017capture}
M.~Bombile and A.~Billard, ``Capture-point based balance and reactive
  omnidirectional walking controller,'' in \emph{IEEE RAS International
  Conference on Humanoid Robots}, no. EPFL-CONF-231920, 2017.

\bibitem{Saab2013}
L.~Saab, O.~E. Ramos, F.~Keith, N.~Mansard, P.~Sou\`{e}res, and J.~Y. Fourquet,
  ``Dynamic whole-body motion generation under rigid contacts and other
  unilateral constraints,'' \emph{IEEE Transactions on Robotics}, vol.~29,
  no.~2, pp. 346--362, April 2013.

\bibitem{Ott2011}
C.~Ott, M.~A. Roa, and G.~Hirzinger, ``Posture and balance control for biped
  robots based on contact force optimization,'' in \emph{2011 11th IEEE-RAS
  International Conference on Humanoid Robots}, pp. 26--33.

\bibitem{Stephens2010}
B.~J. Stephens and C.~G. Atkeson, ``Dynamic balance force control for compliant
  humanoid robots,'' in \emph{2010 IEEE/RSJ International Conference on
  Intelligent Robots and Systems}, pp. 1248--1255.

\bibitem{Herzog2014}
A.~Herzog, L.~Righetti, F.~Grimminger, P.~Pastor, and S.~Schaal, ``Balancing
  experiments on a torque-controlled humanoid with hierarchical inverse
  dynamics,'' in \emph{2014 IEEE/RSJ International Conference on Intelligent
  Robots and Systems}, Sept 2014, pp. 981--988.

\bibitem{lee2012}
S.-H. Lee and A.~Goswami, ``A momentum-based balance controller for humanoid
  robots on non-level and non-stationary ground,'' \emph{Autonomous Robots},
  vol.~33, no.~4, pp. 399--414, Nov 2012.

\bibitem{Nava2016}
G.~Nava, F.~Romano, F.~Nori, and D.~Pucci, ``Stability analysis and design of
  momentum-based controllers for humanoid robots,'' in \emph{2016 IEEE/RSJ
  International Conference on Intelligent Robots and Systems (IROS)}, Oct 2016,
  pp. 680--687.

\bibitem{highlyDynamic}
D.~Pucci, F.~Romano, S.~Traversaro, and F.~Nori, ``Highly dynamic balancing via
  force control,'' in \emph{2016 IEEE-RAS 16th International Conference on
  Humanoid Robots (Humanoids)}, pp. 141--141.

\bibitem{Kajita2003}
S.~Kajita, F.~Kanehiro, K.~Kaneko, K.~Fujiwara, K.~Harada, K.~Yokoi, and
  H.~Hirukawa, ``Biped walking pattern generation by using preview control of
  zero-moment point,'' in \emph{2003 IEEE International Conference on Robotics
  and Automation}, Sept, pp. 1620--1626 vol.2.

\bibitem{Nataleeaaq1026}
L.~Natale, C.~Bartolozzi, D.~Pucci, A.~Wykowska, and G.~Metta, ``icub: The
  not-yet-finished story of building a robot child,'' \emph{Science Robotics},
  vol.~2, no.~13, 2017.

\bibitem{pucci2013nonlinear}
D.~Pucci, L.~Marchetti, and P.~Morin, ``Nonlinear control of unicycle-like
  robots for person following,'' in \emph{Intelligent Robots and Systems
  (IROS), 2013 IEEE/RSJ International Conference on}, pp. 3406--3411.

\bibitem{vukobratovic2004zero}
M.~Vukobratovi{\'c} and B.~Borovac, ``Zero-moment point: thirty five years of
  its life,'' \emph{International journal of humanoid robotics}, vol.~1,
  no.~01, pp. 157--173, 2004.

\bibitem{traversaro2017}
\BIBentryALTinterwordspacing
S.~Traversaro, D.~Pucci, and F.~Nori, ``A unified view of the equations of
  motion used for control design of humanoid robots,'' \emph{On line}, 2017.
  [Online]. Available:
  \url{https://traversaro.github.io/preprints/changebase.pdf}
\BIBentrySTDinterwordspacing

\bibitem{Marsden2010}
J.~E. Marsden and T.~S. Ratiu, \emph{Introduction to Mechanics and Symmetry: A
  Basic Exposition of Classical Mechanical Systems}.\hskip 1em plus 0.5em minus
  0.4em\relax Springer Publishing Company, Incorporated, 2010.

\bibitem{michalska1989receding}
H.~Michalska and D.~Q. Mayne, ``Receding horizon control of nonlinear
  systems,'' in \emph{Decision and Control, 1989., Proceedings of the 28th IEEE
  Conference on}.\hskip 1em plus 0.5em minus 0.4em\relax IEEE, 1989, pp.
  107--108.

\bibitem{Bock84}
H.~Bock and K.~Plitt, ``A multiple shooting algorithm for direct solution of
  optimal control problems*,'' \emph{IFAC Proceedings Volumes}, vol.~17, no.~2,
  pp. 1603 -- 1608, 9th IFAC World Congress: A Bridge Between Control Science
  and Technology, Budapest, Hungary, 2-6 July 1984.

\bibitem{diehl2006fast}
M.~Diehl, H.~G. Bock, H.~Diedam, and P.-B. Wieber, ``Fast direct multiple
  shooting algorithms for optimal robot control,'' in \emph{Fast motions in
  biomechanics and robotics}.\hskip 1em plus 0.5em minus 0.4em\relax Springer
  Berlin Heidelberg, 2006, pp. 65--93.

\bibitem{libiDynTree}
{\relax Dynamic Interaction Control Lab - Istituto Italiano di Tecnologia},
  ``i{D}yn{T}ree library,'' https://github.com/robotology/idyntree, 2016.

\bibitem{IPOpt2006}
A.~W{\"a}chter and L.~T. Biegler, ``On the implementation of an interior-point
  filter line-search algorithm for large-scale nonlinear programming,''
  \emph{Mathematical Programming}, vol. 106, no.~1, pp. 25--57.

\bibitem{olfati2001nonlinear}
R.~Olfati-Saber, ``Nonlinear control of underactuated mechanical systems with
  application to robotics and aerospace vehicles,'' Ph.D. dissertation,
  Massachusetts Institute of Technology, 2001.

\bibitem{Ferreau2014}
H.~Ferreau, C.~Kirches, A.~Potschka, H.~Bock, and M.~Diehl, ``{qpOASES}: A
  parametric active-set algorithm for quadratic programming,''
  \emph{Mathematical Programming Computation}, vol.~6, no.~4, pp. 327--363,
  2014.

\bibitem{hu2016walking}
Y.~Hu, J.~Eljaik, K.~Stein, F.~Nori, and K.~Mombaur, ``Walking of the icub
  humanoid robot in different scenarios: implementation and performance
  analysis,'' in \emph{Humanoid Robots (Humanoids), 2016 IEEE-RAS 16th
  International Conference on}, pp. 690--696.

\end{thebibliography}
